\documentclass[a4paper,11pt]{article}
\title{Feature-Weighted Linear Stacking}
\author{Joseph Sill$^{1}$, Gabor Takacs$^2$, Lester Mackey$^3$, and David Lin$^4$ \\
\date{}
\mbox{}\\
$^1$\tiny{Analytics Consultant (joe\_sill@yahoo.com) } \\
$^2$\tiny{Szechenyi Istvan University, Gyor, Hungary (gtakacs@sze.hu)} \\
$^3$\tiny{University of California, Berkeley (lmackey@cs.berkeley.edu)} \\
$^4$\tiny{david.yang.lin@gmail.com} \\
}
\usepackage{epsfig}
\usepackage{amssymb}
\usepackage{hyperref}
\usepackage{breakurl}

\begin{document}
\maketitle
\begin{abstract} Ensemble methods, such as stacking, are designed to boost predictive accuracy by blending the predictions of multiple machine learning models. Recent work has shown that the use of meta-features, additional inputs describing each example in a dataset, can boost the performance of ensemble methods, but the greatest reported gains have come from nonlinear procedures requiring significant tuning and training time.  Here, we present a linear technique, Feature-Weighted Linear Stacking (FWLS), that incorporates meta-features for improved accuracy while retaining the well-known virtues of linear regression regarding speed, stability, and interpretability.  FWLS combines model predictions linearly using coefficients that are themselves linear functions of meta-features.  This technique was a key facet of the solution of the second place team in the recently concluded Netflix Prize competition.  Significant increases in accuracy over standard linear stacking are demonstrated on the Netflix Prize collaborative filtering dataset.
\end{abstract}

\section{Introduction}

``Stacking'' is a technique in which the predictions of a collection of models are given as inputs to a second-level learning algorithm.  This second-level algorithm is trained to combine the model predictions optimally to form a final set of predictions. Many machine learning practitioners have had success using stacking and related techniques to boost prediction accuracy beyond the level obtained by any of the individual models. In some contexts, stacking is also referred to as blending, and we will use the terms interchangeably here. Since its introduction \cite{wolpert:1992:sg}, modellers have employed stacking successfuly on a wide variety of problems, including chemometrics \cite{Chemo_2007}, spam filtering \cite{Sakkis01}, and large collections of datasets drawn from the UCI Machine learning repository \cite{Issues_In_Stacked,mach:Dzeroski+Zenko:2004}. One prominent recent example of the power of model blending was the Netflix Prize\footnote{http://www.netflixprize.com} collaborative filtering competition. The team \emph{BellKor's Pragmatic Chaos} won the \$1 million prize using a blend of hundreds of different models \cite{BigChaos_Grand,BellKor_Grand,PragmaticTheory_Grand}. Indeed, the winning solution was a blend at multiple levels, i.e., a blend of blends.

Intuition suggests that the reliability of a model may vary as a function of the conditions in which it is used. For instance, in a collaborative filtering context where we wish to predict the preferences of customers for various products, the amount of data collected may vary significantly depending on which customer or which product is under consideration. Model A may be more reliable than model B for users who have rated many products, but model B may outperform model A for users who have only rated a few products. In an attempt to capitalize on this intuition, many researchers have developed approaches that attempt to improve the accuracy of stacked regression by adapting the blending on the basis of side information. Such an additional source of information, like the number of products rated by a user or the number of days since a product was released, is often referred to as a ``meta-feature,'' and we will use that terminology here. 

Unsurprisingly, linear regression is the most common learning algorithm used in stacked regression. The many virtues of linear models are well known to modellers. The computational cost involved in fitting such models (via the solution of a linear system) is usually modest and always predictable. They typically require a minimum of tuning. The transparency of the functional form lends itself naturally to interpretation. At a minimum, linear models are often an obvious initial attempt against which the performance of more complex models is benchmarked. Unfortunately, linear models do not (at first glance) appear to be well suited to capitalize on meta-features. If we simply merge a list of meta-features with a list of models to form one overall list of independent variables to be linearly combined by a blending algorithm, then the resulting functional form does not appear to capture the intuition that the relative emphasis given the predictions of various models should depend on the meta-features, since the coefficient associated with each model is constant and unaffected by the values of the meta-features. 

Previous work has indeed suggested that nonlinear, iteratively trained models are needed to make good use of meta-features for blending. The winning Netflix Prize submission of \emph{BellKor's Pragmatic Chaos} is a complex blend of many sub-blends, and many of the sub-blends use blending techniques which incorporate meta-features. The number of user and movie ratings, the number of items the user rated on a particular day, the date to be predicted, and various internal parameters extracted from some of the recommendation models were all used within the overall blend. In almost all cases, the algorithms used for the sub-blends incorporating meta-features were nonlinear and iterative, i.e., either a neural network or a gradient-boosted decision tree.

In \cite{Xinlong_Bao_Thesis}, a system called STREAM (Stacking Recommendation Engines with Additional Meta-Features) which blends recommendation models is presented. Eight meta-features are tested, but the results showed that most of the benefit came from using the number of user ratings and the number of item ratings, which were also two of the most commonly used meta-features by \emph{BellKor's Pragmatic Chaos}. Linear regression, model trees, and bagged model trees are used as blending algorithms with bagged model trees yielding the best results. Linear regression was the least successful of the approaches. 

Collaborative filtering is not the only application area where the use of meta-features or other dynamic approaches to model blending has been attempted. In a classification problem context \cite{mach:Dzeroski+Zenko:2004}, Dzeroski and Zenko attempt to augment a linear regression stacking algorithm by meta-features such as the entropy of the predicted class probabilities, although they found that it yielded limited benefit on a suite of tasks from the UC Irvine machine learning repository. An approach which does not use meta-features per se but which does employ an adaptive approach to blending is described by Puuronen, Terziyan, and Tsymbal \cite{Puuronen}. They present a blending algorithm based on weighted nearest neighbors which changes the weightings assigned to the models depending on estimates of the accuracies of the models within particular subareas of the input space.  

Thus, a survey of the pre-existing literature suggests that nonparametric or iterative nonlinear approaches are usually required in order to make good use of meta-features when blending. The method presented in this paper, however, can capitalize on meta-features while being fit via linear regression techniques. The method does not simply add meta-features as additional inputs to be regressed against. It parametrizes the coefficients associated with the models as linear functions of the meta-features. Thus, the technique has all the familiar speed, stability, and interpretability advantages associated with linear regression while still yielding a significant accuracy boost. The blending approach was an important part of the solution submitted by \emph{The Ensemble}, the team which finished in second place in the Netflix Prize competition.

\section{Feature-Weighted Linear Stacking}

\subsection{Algorithm}

Let $\mathcal{X}$ represent the space of inputs and $g_1, g_2, \cdots, g_L$ denote the learned prediction functions of $L$ machine learning models with $g_i: \mathcal{X} \rightarrow \mathbb{R}, \forall i$.  In addition, let $f_1,f_2, \cdots, f_M$ represent a collection of $M$ meta-feature functions to be used in blending. Each meta-feature function $f_i$ maps each datapoint $x \in \mathcal{X}$ to its corresponding meta-feature $f_i(x) \in \mathbb{R}$.  Standard linear regression stacking \cite{Breiman} seeks a blended prediction function $b$ of the form 
 
\begin{equation}
b(x) = \sum_{i} w_i g_i(x), \forall x \in \mathcal{X}
\label{eq:standardStacking}
\end{equation}
where each learned model weight, $w_i$, is a constant in $\mathbb{R}$.
 
Feature-weighted linear stacking (FWLS) instead models the blending weights $w_i$ as linear functions of the meta-features, i.e.
\begin{equation}
w_i(x) = \sum_{j} v_{ij} f_j(x), \forall x \in \mathcal{X}
\end{equation}
for learned weights $v_{ij} \in \mathbb{R}$.
Under this assumption, Eq.~\ref{eq:standardStacking} can be rewritten as 
\begin{equation}
b(x) = \sum_{i,j} v_{ij}f_j(x) g_i(x), \forall x \in \mathcal{X}
\end{equation}
yielding the following FWLS optimization problem:
\begin{equation}
\min_v \sum_{x\in \tilde{\mathcal{X}}} \sum_{i,j} (v_{ij}f_j(x) g_i(x) - y(x))^2.
\end{equation}
where $y(x)$ is the target prediction for datapoint $x$ and $\tilde{\mathcal{X}}$ is the subset of $\mathcal{X}$ used to train the stacking parameters.

We thereby obtain an expression for $b$ which is linear in the free parameters $v_{ij}$, and we can use a single linear regression to estimate those parameters. The independent variables of the regression (i.e., the ``inputs'', in machine learning parlance) are the $ML$ products $f_j(x) g_i(x)$ of meta-feature function and model predictor evaluated at each $x \in \tilde{\mathcal{X}}$. 
\begin{center}
\begin{figure}[htp]
\includegraphics[width=\textwidth]{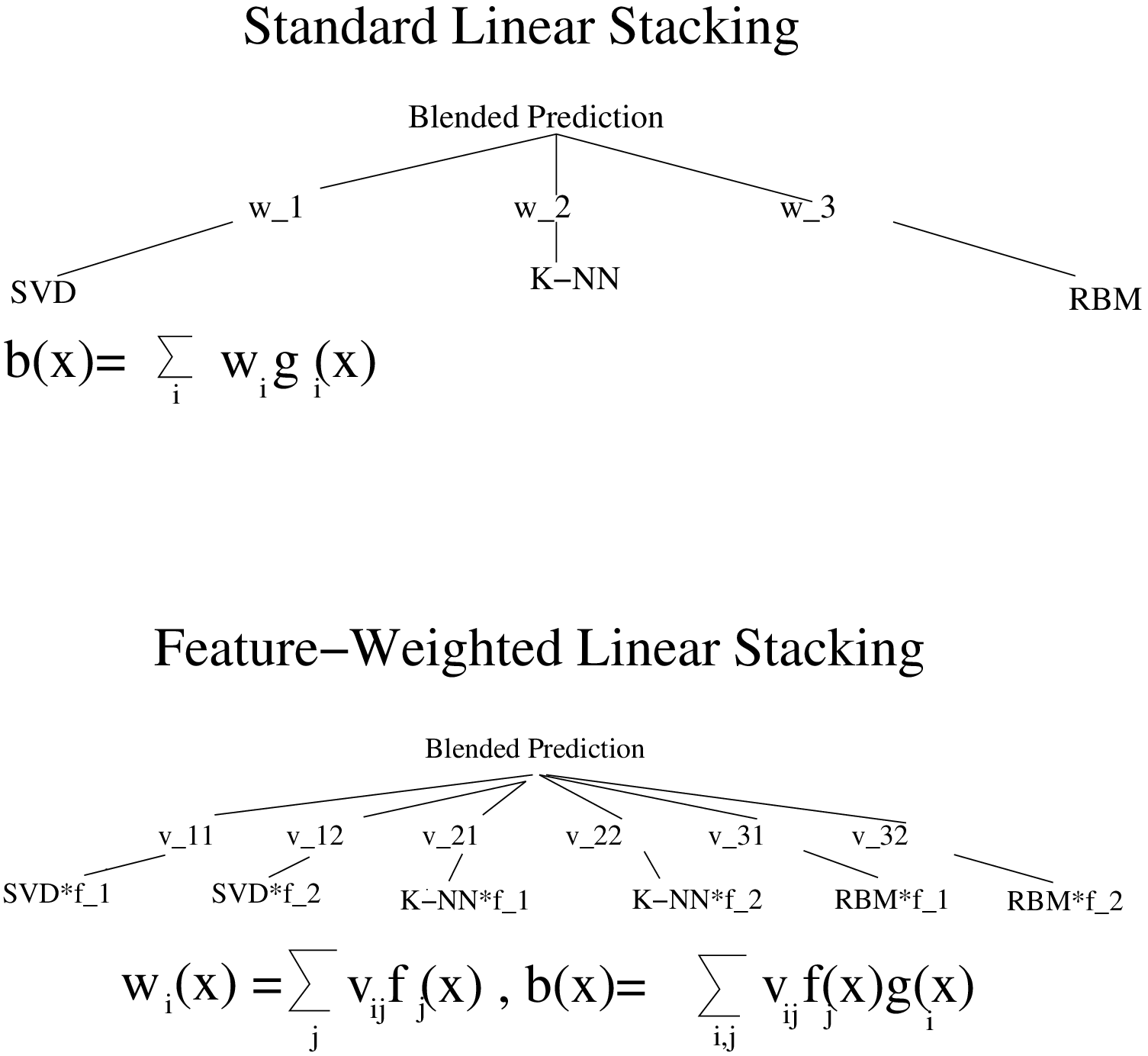}
\caption{FWLS forms a linear combination of products of model outputs and meta-features}
\label{fig:fwlsFig1}
\end{figure}
\end{center}

Figure~\ref{fig:fwlsFig1} shows a graphical interpretation of  FWLS. The outputs of individual models $g_{i}$ are represented as SVD, K-NN, and RBM in the figure. These acronyms represent common collaborative filtering algorithms which will be described in the next section.  While it is helpful conceptually to think of a linear combination of models where the coefficients of the combination vary as a function of meta-features, the figure portrays the alternative but equivalent interpretation corresponding to equation 3 and corresponding to a concrete software implementation, i.e., a regression against all possible two-way products of models and meta-features.

An alternate interpretation of FWLS is to view it as a kind of bipartite quadratic regression against a set of independent variables consisting of the models and the meta-features. We obtain the FWLS form by starting with a full quadratic regression and dropping the terms resulting from interacting the models with themselves and each other, as well as the terms resulting from interacting the meta-features with themselves and each other.  

It is important that the dataset collected for the stacked regression consists of out-of-sample model predictions. In other words, to obtain the prediction of a model on a particular data point, the model parameters should have been fitted on a training set which does not include that data point. This is normally achieved by via $K$-fold cross-validation. The training data is split into $K$ subsets and $K$ versions of the model are trained, each on a version of the data with a different subset removed. Model predictions for the $k$th subset are generated from the version of the model whose training set did not include that subset. Occasionally, however, the data distribution the models are to be tested on is not the same as the distribution from which the training data was drawn, in which case the blending procedure may differ \footnote[1]{In this situation, if there is a small subset $S$ of the training data drawn from the same distribution as the test data, then that subset can be removed from the training data and used instead to fit the stacking linear regression model, since the blend should be optimized with respect to the test distribution. In order to maximize accuracy to the fullest, the models can then be retrained on the full training set, including $S$, and the blending function obtained from the stacking linear regression can be used to blend the predictions of the retrained models. Although this procedure may be difficult to justify with full rigor, it can work well in practice. This was the approach commonly taken during the Netflix Prize competition, which will be described in more detail in section 3.}.     

It is reasonable to assume that there should be a constant component to the $w_i$ as well as a component which varies with the $f_j$. In the experiments shown in the next section of the paper, we do indeed allow for a constant component of the weights. We can represent this within the above notation by defining $f_0$ to be a special meta-feature which always takes the value $1$. Similarly, one might expect the $f_j$ to add some modest amount of value when included as inputs to the regression on their own (i.e., without interacting them with the $g_i$). The above notation can be understood to cover this case by including a special, constant model $g_0$ which always takes the value $1$.

The number of estimated parameters, $ML$, can be substantial when blending a large collection of models using a long-list of meta-features. Ridge regression (a.k.a. Tikhonov regularization) can be used to combat overfitting in these cases, such as the experimental results presented later in this paper.

It should be noted that there was prior work in which the FWLS functional form was employed on a small scale, but with important differences. In their 2008 Netflix Progress Prize paper \cite{BellKor_Progress_2008}, Bell, Koren and Volinsky make use of two meta-features (number of user and number of movie ratings) within a linear model in a construction which is similar to the formulation we present here, although their approach also includes coefficients which are specific to each movie. Perhaps more importantly, their approach employs stochastic gradient descent in order to fit the blending parameters rather than the classic linear-system-solution approach to regression which we advocate here. The results of their specific blending effort appear to play a minor role in their overall blend. 

\subsection{Implementation Details}

Let $N$ be the number of data points used in the stacking regression, and let $A$ be the $N\times ML$ matrix with elements $A_{n,(i+L(j-1))} = f_j(x_n) g_i(x_n)$, where $x_n$ is the input vector for the $n$th data point in $\tilde{\mathcal{X}}$.  Performing a linear regression with Tikhonov regularization amounts to solving the system 
\begin{equation}
 (A^{T}A + \lambda I)v = A^{T}y
 \label{eq:linsystem}
\end{equation}
where $y$ represents the vector of target outputs for the $N$ data points, and $\lambda$ is a given regularization parameter.

The time complexity of FWLS is $O(NM^2L^2 + M^3L^3)$, where the first term corresponds to the cost of computing $A^{T}A$ and the second term corresponds to the solution of the linear system. In practice, $N$ is normally much larger than $ML$ and almost all of the computational cost comes from computing $A^{T}A$. For many realistic scenarios, this computation can be completed quickly. For instance, for the parameters $N=162,731$, $M=26$,and $L=10$, the entire regression finishes in 1 minutes and 35 seconds on a single core of a 1.8Ghz Intel T7100 processor. For very large problems in which hundreds of models are blended using dozens of meta-features, however, the computational cost can be significant. Fortunately, the computation of $A^{T}A$ naturally lends itself to parallelization (e.g. by multithreading) so multiple cores can easily be capitalized on. In such large-problem scenarios, a naive implementation in which the entire $A$ matrix is represented simultaneously in memory could run into memory constraint difficulties. It is straightforward, however, to implement an approach which calculates the entries of $A^{T}A$ directly without ever forming the entirety of $A$ in memory at the same time. This approach requires $O(M^{2}L^{2})$ memory and can be executed by iterating only once over the training data.

In the dynamic setting, where models or meta-features are gradually added to a blend over time, significant computation is saved by serializing previously computed matrices $A^{T}A$ and $A^{T} y$ to disk. When a new model or meta-feature arrives, the previous results can be reloaded, and only the new entries (i.e., those involving the new model or meta-feature) of $A^{T}A$ and $A^{T}y$ need to be computed. This reduces the computational cost of adding a new model to $O(NM^{2}L + M^3L^3)$ and the cost of adding a new meta-feature to $O(NML^{2} + M^3L^3)$, assuming that linear system is solved from scratch. A faster approach is to use the Sherman-Morrison formula \cite{Sherman49} for updating the inverse of a matrix, in which case the second terms in the two preceding expressions can be improved to $M^3L^2$ and $M^2L^3$, respectively. Similarly, adding a single new data point to an existing, saved blend is $O(M^3L^3)$ if the linear system is solved from the beginning every time and only $O(M^2L^2)$ if the Sherman-Morrison formula for the inverse is employed.  

\section{Experiments}

\subsection{Netflix Prize Overview}
The Netflix Prize dataset is a collection of ratings (1 through 5 stars) submitted by customers of the DVD rental company Netflix.  Each rating indicates how much a customer liked a particular movie seen in the past. There are $480,189$ users, $17,770$ movies, and $100,480,507$ movie/user pairs for which the rating is supplied. The date on which the rating was made is also included. A ``qualifying set'' of 2,817,131 movie/user pairs was constructed where the rating the user made was not supplied to competitors\footnote{Now that the contest is over, those ratings are available, along with the rest of the dataset, at the UC Irvine Machine Learning Repository\cite{Asuncion+Newman:2007}.}. Competitors were asked to submit rating predictions for the qualifying set. The qualifying set was derived from a larger set of $4,225,526$ data points formed by collecting the 9 most recent ratings from each user. This larger set was randomly split into 3 subsets of equal size: the probe set, the quiz set, and the test set. The probe set (including the ratings) was included in the training data. The quiz set and the test set together formed the qualifying set, although competitors were not told which of the 2.8 million data points were in one set or the other. The quiz set had virtually no bearing on the official outcome of the competition, but the accuracy of teams' predictions on the quiz set was reported on a publicly viewable leaderboard during the competition. The prediction accuracy teams achieved on the test set determined who won the prize. 

In order to qualify for the prize, a team had to improve upon the accuracy of Netflix's pre-existing algorithm, Cinematch, by at least 10\% on the test set in terms of root mean squared error (RMSE). Since Cinematch's test RMSE was 0.9525, an improvement of 0.0001 in terms of raw RMSE closely corresponded to a 1 basis point (0.01\%) percentage improvement.  Test set scores were unknown to anyone other than Netflix during the competition to ensure that the test set served as a truly out-of-sample evaluation of the submitted solutions. Prior to the awarding of the \$1 million grand prize, there were also two \$50,000 ``Progress Prizes'' awarded in the fall of 2007 and 2008 to the teams with the best scores at that point in the competition.

An overview of the techniques used to win the prizes is presented in the papers written by the prize winners \cite{BigChaos_Grand,BellKor_Grand,PragmaticTheory_Grand}. We briefly summarize a few of the main techniques here in order to provide background for the meta-features selected. Perhaps the most important class of algorithms proved to be matrix factorization techniques, sometimes referred to as by SVD (singular value decomposition) techniques. See \cite{Takacs2009} for an overview of these techniques.  This simplest version of this approach represents each user and each movie as a vector of $F$ real numbers. The predicted rating is given by the dot product of the user vector with the movie vector. The user and movie vector parameters are minimized on the training data, although regularization is typically employed as well. This is called a matrix factorization approach because the $U$ by $M$ rating matrix of all possible (user,movie) pairs is approximated by a low-rank matrix which is the product of a $U$ by $F$ matrix of user parameters and the transpose of an $M$ by $F$ matrix of movie parameters. More sophisticated versions add various additional parameters, such as means for each user and movie and parameters which model time effects, including some which model single-day effects. NSVD1 is an important variation on SVD first proposed by Paterek \cite{Paterek}. This model represents a user as the sum of a set of vectors corresponding to the movies the user has seen, where these vectors are distinct from the vectors comprising the aforementioned $M$ by $F$ movie matrix.  

Perhaps the second most prominent class of algorithms used in the prize-winning solution are the nearest neighbor models, which have a longer and more widespread academic and commerical collaborative filtering history. Nearest neighbor (K-NN) algorithms use a measure of similarity between the movie to be predicted and the movies the user has already rated in order to generate a prediction.  The most common similarity measure involves computing the correlation between the ratings two movies received from the same set of users \cite{SIGIR'99*230}, although other similarity measures were also employed.  Standard approaches take a weighted average of the user's ratings on the $K$ most similar movies, where the weighting is a function of the similarity level. Many variations of this approach were also implemented (e.g. \cite{conf/icdm/BellK07,conf/kdd/Koren08}). There is also a user-based version of nearest neighbors where the ratings which correlated users gave the movie to be predicted are used to generate a prediction, but this proved to be much less useful on the Netflix Prize dataset. Restricted Boltzmann machines (RBMs) \cite{conf/icml/SalakhutdinovMH07}, a kind of stochastic recurrent neural network, are a third major class of algorithms.   

Many algorithms involved a kind of preprocessing known amongst the Netflix Prize community as the removal of global effects, which was largely pioneered by Bell, Koren, and Volinsky \cite{BellKor_Progress_2007} . Global effects are predictions which can be made without simultaneous knowledge of the specific identities of the both the user and the movie. The two simplest examples of global effects are the average rating of the user and the average rating of the movie, although many others have also been found. Global effects are estimated in succession, so the average rating of the movie, for instance, might be estimated on the residual of the rating after subtracting out the average rating of the user.

\subsection{Results}

We present the results of FWLS on 119 of the models of one of the leading teams in the Netflix Prize competition, \emph{Grand Prize Team}. It should be noted that the team name was only aspirational in nature, as the team did not ultimately win the grand prize. However, it did form half of a larger coalition known as \emph{The Ensemble}, which tied \emph{BellKor's Pragmatic Chaos} in terms of test RMSE and finished in second only because its best submission was made 20 minutes after the best submission of \emph{BellKor's Pragmatic Chaos}.
 
Since the probe set was statistically representative of the test set, it was standard practice among Netflix competitors to use the probe data for the sake of fitting a blend, and we followed this procedure here. Two versions of each of the 119 models were trained, with the first version being fitted a training set with the probe set removed and the second version being fitted on a training set including the probe set. The first version of the models was used to generate probe set predictions, and the FWLS regressions were performed using the probe set.The final blending parameters (i.e., the $v_{ij}$) used to generate the qualifying set predictions were obtained by fitting on the entire probe set and then using those parameters to blend the second version of the models, those that were fitted on the training set with the probe set included.

Note that when parameters are chosen to minimize squared error on the probe set, reductions in probe set RMSE will not be entirely reflective of reductions in test set RMSE. In evaluating our methods, we addressed this issue by computing out-of-sample (OOS) probe set RMSE based on 10-fold cross validation.
Ten blends were fit, each on a version of the probe set with a different 10\% removed, and out-of-sample predictions were generated from each blend on the portion of the data on which it was not fit.  

\begin{table}
\caption {Meta-Features used for Netflix Prize model blending}
\small{
\begin{tabular}{|l| p{12cm} |}
\hline
1 & A constant 1 voting feature (this allows the original predictors to be regressed against in addition to their interaction with the voting features) \\
\hline
2 & A binary variable indicating whether the user rated more than 3 movies on this particular date \\
\hline
3 & The log of the number of times the movie has been rated \\
\hline
4 & The log of the number of distinct dates on which a user has rated movies \\
\hline
5 & A bayesian estimate of the mean rating of the movie after having subtracted out the user's bayesian-estimated mean \\
\hline
6 & The log of the number of user ratings \\
\hline
7 & The mean rating of the user, shrunk in a standard bayesian way towards the mean over all users of the simple averages of the users \\
\hline
8 & The norm of the SVD factor vector of the user from a 10-factor SVD trained on the residuals of global effects \\
\hline
9 & The norm of the SVD factor vector of the movie from a 10-factor SVD trained on the residuals of global effects  \\
\hline
10 & The log of the sum of the positive correlations of movies the user has already rated with the movie to be predicted \\
\hline
11 & The standard deviation of the prediction of a 60-factor ordinal SVD  \\
\hline
12 & Log of the average number of user ratings for those users who rated the movie \\
\hline
13 & The log of the standard deviation of the dates on which the movie was rated. Multiple ratings on the same date are represented multiple times in this calculation \\ 
\hline
14 & The percentage of the correlation sum in feature 10 accounted for by the top 20 percent of the most correlated movies the user has rated. \\
\hline
15 & The standard deviation of the date-specific user means from a model which has separate user means (a.k.a biases) for each date \\
\hline
16 & The standard deviation of the user ratings \\
\hline
17 & The standard deviation of the movie ratings \\
\hline
18 & The log of (rating date - first user rating date + 1) \\
\hline
19 & The log of the number of user ratings on the date + 1 \\
\hline
20 & The maximum correlation of the movie with any other movie, regardless of whether the other movies have been rated by the user or not \\ 
\hline
21 & Feature 3 times Feature 6, i.e., the log of the number of user ratings times the log of the number of movie ratings \\
\hline
22 & Among pairs of users who rated the movie, the average overlap in the sets of movies the two users rated, where overlap is defined as the percentage of movies in the smaller of the two sets which are also in the larger of the two sets. \\
\hline
23 & The percentage of ratings of the movie which were the only rating of the day for the user \\
\hline
24 & The (regularized) average number of movie ratings for the movies rated by the user. \\
\hline
25 &  First, a movie-movie matrix was created with entries containing the probability that the pair of movies was rated on the same day conditional on a user having rated both movies. Then, for each movie, the correlation between this probability vector over all movies and the vector of ratings correlations with all movies was computed\\
\hline
\end{tabular}
}
\label{tab:metafeats}
\end{table}

  Based on 00S probe set RMSE, we found a list of 24 meta-features which proved to be helpful. A description of these meta-features is shown in Table~\ref{tab:metafeats}. A blend which uses only meta-feature 1 (the constant meta-feature which always takes the value $1$) is equivalent to standard linear regression stacking (this trivial meta-feature is not included when arriving at the count of 24 useful meta-features). The creation of useful meta-features is an art which is guided by a detailed understanding of the characteristics of the models to be blended and an intuition about conditions under which certain models might merit greater emphasis. For the sake of brevity, we will only discuss the reasoning behind a handful of the 24 meta-features. 

Meta-features 3 and 6 (the log of the number of user and movie ratings, respectively) are the most commonly used meta-features in the previously existing literature. The intuition behind their usage is fairly clear, i.e., the relative accuracy of models may depend on how much information we have (i.e., how many ratings have been collected) regarding particular users and movies. The reasoning behind meta-features 12 and 24 is similar. The information we have about a particular movie depends not only on the number of users who rated it but also on whether or not those users have rated many movies. An analogous statement can be made if we switch users and movies in the previous statement.  

As mentioned in the previous section, many nearest-neighbor-based techniques rely on the estimated correlations between pairs of movies. Meta-features 10 and 14 attempt to characterize the correlation information available regarding the particular (user,movie) pair to be predicted. If meta-feature 10 is large, then the user has rated many movies which have a significant correlation with the movie to be predicted, which should bode well for neighborhood-based techniques. Meta-feature 14 measures whether the total represented in meta-feature 10 is concentrated in just a few very highly correlated movies or whether it is distributed across a larger set of movies. 

Many models attempt to capture effects which vary over time, and there are several meta-features designed to reflect this. Some models associate a mean rating (a.k.a. a bias) with each distinct date on which the user rated movies \cite{conf/kdd/Koren09}. If the date-specific means for certain users vary a great deal from day to day, then one might guess that the models which capture these effects should be given more emphasis in those cases. Meta-feature 15 is motivated by this observation. Meta-feature 4 and 13 are also designed to suggest the importance of time effects by measuring, in different ways, how dispersed in time the ratings are. 

It is possible to implement a version of an SVD algorithm which produces a probability distribution over the 5 possible ratings and hence a standard deviation indicating the uncertainty around the predicted rating. The specific approach used to derive meta-feature 11 is described in \cite{Ordinal}, although an alternate technique which builds 4 separate models for the probability that the rating is less than or equal to $r, \: 1 \le r \le 4$ is described in \cite{PragmaticTheory_Grand}. A high standard deviation may reasonably be interpreted as low confidence in the model's prediction, so it is not surprising that this meta-feature is useful for blending SVD algorithms with other algorithms. 

\begin{table}

\caption {RMSEs Using Cumulative Meta-Feature Sets }
\begin{center}
\begin{tabular}{|c|c|c|}
\hline
Meta-Feature & Probe CV RMSE & Test RMSE \\
\hline
1 & 0.869889 & 0.863377 \\
\hline 
2 & 0.869513 & 0.862914 \\
\hline
3 & 0.869092 & 0.862508 \\
\hline
4 & 0.868734 & 0.862237 \\
\hline
5 & 0.868612 & 0.862210 \\
\hline
6 & 0.868348 & 0.862060 \\
\hline
7 & 0.868271 & 0.862028 \\
\hline
8 & 0.868238 &  0.861994 \\
\hline
9 & 0.868227 & 0.861978 \\
\hline
10 & 0.868163 & 0.861935 \\
\hline
11 & 0.868023 & 0.861834 \\
\hline
12 & 0.867967 & 0.861786 \\
\hline
13 & 0.867890 & 0.861695 \\
\hline
14 & 0.867861 & 0.861657 \\
\hline
15 & 0.867846 & 0.861580 \\
\hline
16 & 0.867773 & 0.861507 \\
\hline
17 & 0.867745 & 0.861532 \\
\hline
18 & 0.867707 & 0.861552 \\
\hline
19 & 0.867636 & 0.861532 \\
\hline
20 & 0.867618 & 0.861494 \\
\hline 
21 & 0.867575 & 0.861504 \\
\hline
22 & 0.867547 & 0.861475 \\
\hline
23 & 0.867543 & 0.861498 \\
\hline
24 & 0.867512 & 0.861456 \\
\hline
25 & 0.867501 & 0.861405 \\
\hline 
\end{tabular}
\end{center}
\end{table}
  
Table 2 shows the cross-validated probe set RMSE results(based on version 1 of the models) and also the RMSE on the test set (based on version 2 of the models). Row $m$ corresponds to the results using meta-features $1$ through $m$, so the difference between RMSEs in rows $m-1$ and $m$ represents the incremental contribution of adding the $m$th meta-feature to the set of meta-features. As was previously mentioned, the test set RMSE was unknown to competitors during the competition, so the set of meta-features was developed without knowing those results. Since the contribution of some meta-features to accuracy on the probe set is modest, it is unsurprising to learn that a few of the meta-features were in fact mildly harmful to the test set RMSE, but in general the cross-validated probe set RMSE is a reasonably reliable indicator of the test set RMSE. The meta-features contribute 23.88 basis points of accuracy on the probe set and 19.72 basis points of accuracy on the test set. It is, of course, debatable whether all meta-features would be used in a commercial context, given that the computational cost of fitting the blend grows quadratically with the number of meta-features. In the context of the Netflix Prize competition, however, every basis point of improvement was precious.

We argued in the introduction that a simple linear regression approach which merely includes the meta-features as additional independent variables to be regressed against is ill-suited as an approach to stacking with meta-features. Our experiments on the Netflix Prize data confirm this suspicion. Including the same 24 meta-features as additional inputs to the regression yields a cross-validated probe set RMSE of 0.868641, i.e., only 1 basis point better than the RMSE obtained without using the meta-features at all.

It is important to note that meta-features were added to the collection one by one after demonstrating an ability to decrease the RMSE obtained with the previous set of meta-features. For this reason, the set of meta-features arrived at is ``path dependent'' in the sense that it depended on the order in which potential new meta-features occurred to the authors and were tested. There were many other meta-features evaluated which are not listed in Table 1 which were highly valuable when used in isolation, in the sense that the RMSE with the meta-feature alone significantly improved upon the RMSE obtained by standard linear regression with no meta-features. However, such meta-features did not improve the RMSE of the blend when using the entire set of meta-features already employed. It is possible that by removing some of the meta-features already in the set and adding in new meta-features under consideration that a superior set of meta-features could have been obtained, but this line of experimentation was generally not pursued.

\section{Discussion}

There are several potential extensions of this work, many of which we plan to pursue and present in a longer paper. In one of the classic papers on stacking \cite{Breiman}, Breiman strongly advocates the use of non-negative weights when using a linear blending model. The results we have presented do not constrain the $v_{ij}$ in any way, but work is underway to evaluate the value of using nonnegativity constraints for FWLS. 

Another line of research to pursue will involve pruning the expanded space of $ML$ model/meta-feature pairs in order to reduce the number of parameters estimated and thereby perhaps improve upon both out-of-sample accuracy and the speed with which the blend can be fit. There are a variety of linear model feature selection and pruning algorithms which may be applicable here. 

It is also likely the case that meta-features discovered in the course of using FWLS would be useful as inputs to nonlinear blenders such as neural networks and trees. Indeed, initial experiments involving neural networks confirm this suspicion and suggest that a second-level blending of a neural network blend and an FWLS blend, both using the same set of meta-features, yields accuracy superior to either individual blend. We plan to present detailed results on this topic in the future.  The speed of FWLS, when used for blending a moderately-sized model collection, allows for quick discovery of useful meta-features which can then be passed on for use with neural networks, trees, and other nonlinear techniques. Thus, even if another blending approach is strongly preferred, FWLS may have value as a mechanism for discovering meta-features. 

The interpretability of linear models is not to be forgotten as one of the additional merits of the approach. FWLS affords the opportunity to examine the effective coefficients associated with each model under various conditions, i.e., various values of the meta-features. Scrutiny of such coefficients may lead to insights regarding the conditions under which the various models are most successful.

Finally, the authors wish to emphasize that FWLS should in principle be applicable to a wide variety of situations in which stacking is employed, so applications to domains other than collaborative filtering will be explored in the future. 

\bibliographystyle{plain}
\bibliography{fwlsPaper}

\begin{thebibliography}{10}

\bibitem{Asuncion+Newman:2007}
A.~Asuncion and D.J. Newman.
\newblock {UCI} machine learning repository, 2007.

\bibitem{Xinlong_Bao_Thesis}
Xinlong Bao.
\newblock Applying machine learning for prediction, recommmendation, and
  integration, August 2009.
\newblock
  \url{http://scholarsarchive.library.oregonstate.edu/jspui/bitstream/1957/125%
49/1/Dissertation_XinlongBao.pdf}.

\bibitem{BellKor_Progress_2007}
Robert Bell, Yehuda Koren, and Chris Volinsky.
\newblock The bellkor solution to the netflix prize, November 2007.
\newblock
  \url{http://www.netflixprize.com/assets/ProgressPrize2007_KorBell.pdf}.

\bibitem{BellKor_Progress_2008}
Robert Bell, Yehuda Koren, and Chris Volinsky.
\newblock The bellkor 2008 solution to the netflix prize, December 2008.
\newblock
  \url{http://www.netflixprize.com/assets/ProgressPrize2008_BellKor.pdf}.

\bibitem{conf/icdm/BellK07}
Robert~M. Bell and Yehuda Koren.
\newblock Scalable collaborative filtering with jointly derived neighborhood
  interpolation weights.
\newblock In {\em ICDM}, pages 43--52. IEEE Computer Society, 2007.

\bibitem{Breiman}
Leo Breiman.
\newblock Stacked regressions.
\newblock {\em Machine Learning}, 24:49--64, 1996.

\bibitem{mach:Dzeroski+Zenko:2004}
Saso D{\^z}eroski and Bernard {\^Z}enko.
\newblock Is combining classifiers with stacking better than selecting the best
  one?
\newblock {\em Machine Learning}, 54(3):255--273, 2004.
\newblock Special Issue: Meta-Learning.

\bibitem{Chemo_2007}
Lu~Xu et. al.
\newblock Mccv stacked regression for model combination and fast spectral
  interval selection in multivariate calibration.
\newblock {\em Chemometrics and Intelligent Laboratory Systems}, 87:226--230,
  2007.

\bibitem{SIGIR'99*230}
Jonathan~L. Herlocker, Joseph~A. Konstan, Al~Borchers, and John Riedl.
\newblock An algorithmic framework for performing collaborative filtering.
\newblock pages 230--237.

\bibitem{conf/kdd/Koren08}
Yehuda Koren.
\newblock Factorization meets the neighborhood: a multifaceted collaborative
  filtering model.
\newblock In Ying Li, Bing Liu, and Sunita Sarawagi, editors, {\em KDD}, pages
  426--434. ACM, 2008.

\bibitem{BellKor_Grand}
Yehuda Koren.
\newblock The bellkor solution to the netflix grand prize, September 2009.
\newblock
  \url{http://www.netflixprize.com/assets/GrandPrize2009_BPC_BellKor.pdf}.

\bibitem{conf/kdd/Koren09}
Yehuda Koren.
\newblock Collaborative filtering with temporal dynamics.
\newblock In John F.~Elder IV, Fran{\c c}oise Fogelman-Souli{\'e}, Peter~A.
  Flach, and Mohammed~Javeed Zaki, editors, {\em KDD}, pages 447--456. ACM,
  2009.

\bibitem{Paterek}
A.~Paterek.
\newblock Improving regularized singular value decomposition for collaborative
  filtering, 2007.
\newblock KDD-Cup and Workshop, ACM press, 2007.

\bibitem{PragmaticTheory_Grand}
Martin Piotte and Martin Chabbert.
\newblock The pragmatic theory solution to the netflix grand prize, September
  2009.
\newblock
  \url{http://www.netflixprize.com/assets/GrandPrize2009_BPC_PragmaticTheory.p%
df}.

\bibitem{Puuronen}
Seppo Puuronen, Vagan Terziyan, and Alexey Tsymbal.
\newblock A dynamic integration algorithm for an ensemble of classifiers.
\newblock {\em Foundations of Intelligent Systems}, 1609:592--600, 1999.

\bibitem{Sakkis01}
Georgios Sakkis, Ion Androutsopoulos, Georgios Paliouras, Vangelis Karkaletsis,
  Constantine~D. Spyropoulos, and Panagiotis Stamatopoulos.
\newblock Stacking classifiers for anti-spam filtering of {E}-mail.
\newblock In Lillian Lee and Donna Harman, editors, {\em Proceedings of
  EMNLP-01, 6th Conference on Empirical Methods in Natural Language
  Processing}, pages 44--50, Pittsburgh, US, 2001. Association for
  Computational Linguistics, Morristown, US.

\bibitem{conf/icml/SalakhutdinovMH07}
Ruslan Salakhutdinov, Andriy Mnih, and Geoffrey~E. Hinton.
\newblock Restricted boltzmann machines for collaborative filtering.
\newblock In Zoubin Ghahramani, editor, {\em ICML}, volume 227 of {\em ACM
  International Conference Proceeding Series}, pages 791--798. ACM, 2007.

\bibitem{Sherman49}
Jack Sherman and Winifred~J. Morrison.
\newblock Adjustment of an inverse matrix corresponding to changes in the
  elements of a given column or a given row of the original matrix.
\newblock {\em Annals of Mathematical Statistics}, 20:621, 1949.

\bibitem{Ordinal}
Joseph Sill.
\newblock Ordinal matrix factorization, September 2009.
\newblock \url{http://www.netflixprize.com/community/viewtopic.php?id=1541}.

\bibitem{Takacs2009}
G{\'a}bor Tak{\'a}cs, Istv{\'a}n Pil{\'a}szy, Botty{\'a}n N{\'e}meth, and
  Domonkos Tikk.
\newblock Scalable collaborative filtering approaches for large recommender
  systems, March 2009.

\bibitem{Issues_In_Stacked}
Kai~Ming Ting and Ian~H. Witten.
\newblock Issues in stacked generalization.
\newblock {\em Journal of Artificial Intelligence Research}, 10:271--289, 1999.

\bibitem{BigChaos_Grand}
Andreas Toscher, Michael Jahrer, and Robert Bell.
\newblock The bigchaos solution to the netflix grand prize, September 2009.
\newblock
  \url{http://www.netflixprize.com/assets/GrandPrize2009_BPC_BigChaos.pdf}.

\bibitem{wolpert:1992:sg}
David~H. Wolpert.
\newblock Stacked generalization.
\newblock {\em Neural {N}etworks}, 5:241--259, 1992.

\end{thebibliography}
\end{document}